\documentclass[11pt]{article}
\usepackage{amsmath, amsfonts, amssymb, amsthm, mathtools}
\usepackage{graphicx}
\usepackage{booktabs}
\usepackage{natbib}
\usepackage{hyperref}
\usepackage[margin=1.25in]{geometry}
\usepackage{microtype}

\theoremstyle{plain}
\newtheorem{theorem}{Theorem}[section]

\newtheorem{corollary}[theorem]{Corollary}

\theoremstyle{remark}

\newcommand{\E}{\mathbb{E}}
\newcommand{\Var}{\operatorname{Var}}
\renewcommand{\P}{\mathbb{P}}
\newcommand{\R}{\mathbb{R}}
\newcommand{\phat}{\widehat{p}}

\newcommand{\cX}{\mathcal{X}}

\begin{document}

\title{PAC-style Bounds on the Conditional Validity\\
       of Conformal Predictors}

\author{Nicolo Colombo\\
        Department of Computer Science\\
        Royal Holloway, University of London\\
        \texttt{nicolo.colombo@rhul.ac.uk}}

\date{}

\maketitle

\begin{abstract}
Conformal prediction sets are guaranteed to be marginally valid over the attribute space, but local coverage can diverge from the user-defined target $1-\alpha$.  
Constructing conditionally valid prediction sets with finite calibration data is impossible for a general data-generating distribution.  
We derive PAC-type probabilistic lower bounds on the input-conditional coverage of any conformal predictor, without assumptions on the data distribution.  
They are based on the Markov and Cantelli inequalities and adapt to both the calibration-marginal and calibration-conditional setups. 
A tighter, model-dependent Cantelli bound depends on the variance of the conditional coverage and is estimable from data using a fully non-asymptotic Hoeffding argument.  
The results suggest conditional-coverage variance as a scalable and theoretically motivated empirical measure of conditional validity.
\end{abstract}

\section{Introduction}
\label{sec:intro}

Learning machines $f$ predict the unknown labels $Y\in \mathcal{Y}$ of given attributes $X \in \cX$.  
Conformal prediction algorithms leverage labelled calibration sets $\{(X_i, Y_i)\}_{i=1}^{n}$ to promote point predictions $f(X) \approx \E(Y \mid X)$ to prediction sets $\Gamma_\alpha(X)$ that contain the true label with lower-bounded probability,
\begin{equation}
\label{prediction set}
    \P_{XYD}\{Y \in \Gamma_\alpha(X)\} \geq 1-\alpha,
\end{equation}
for any user-defined $\alpha \in (0, 1)$.  
Coverage holds under mild exchangeability assumptions on calibration and test samples, which are satisfied when $(X, Y)$ and $(X_i, Y_i) \in D$ are independent draws from the same distribution.  
The set $\Gamma_\alpha(X)$ is marginally valid because the probability in~\eqref{prediction set} averages over the calibration set, the test attribute, and the test response.  For a continuous $X$,
\begin{align}
\label{prediction set marginal}
    \P_{XYD}\{Y\in \Gamma_\alpha(X)\}
    &= \int \P_{YD}\{Y\in \Gamma_\alpha(X) \mid X\} \,\mathrm{d}\P_X \\
    &= \int\!\!\int \P_{Y}\{Y\in \Gamma_\alpha(X) \mid X, D\}
       \,\mathrm{d}\P_X \,\mathrm{d}\P_D. \nonumber
\end{align}
Obtaining distribution-free input-conditional guarantees for the conditional factors $P = p(X) = \P_{YD}\{Y\in \Gamma_\alpha(X)\mid X\}$ or $P_D = p_D(X) = \P_{Y}\{Y\in \Gamma_\alpha(X)\mid X, D\}$ is impossible for finite calibration sets \citep{lei2012distribution, vovk2012conditional}.
Conformal prediction research on input-conditional coverage has used approximate definitions of conditional validity and produced lower bounds that hold asymptotically or under regularity assumptions on $p(X)$ \citep{gibbs2025conformal, guan2023localized, izbicki2019flexible}.

Guarantees on the input-marginal, calibration-conditional coverage $\P_{XY}\{Y\in \Gamma_\alpha(X)\mid D\}$ have been obtained by \citet{vovk2012conditional} and \citet{duchi2025few} using PAC-style two-parameter lower bounds, in which an additional user-defined parameter controls the algorithm's success probability, 
\begin{equation}
\label{calibration conditional}
    \P_D\bigl[\P_{XY}\{Y\in \Gamma_\alpha(X)\mid D\}
    \geq 1-\alpha - \gamma \bigr] \geq 1 - e^{-2n\gamma^2}.
\end{equation}
Extending the PAC-style approach to the input-conditional setup is our main contribution.

Theorems~\ref{theorem lower} and~\ref{theorem variance exact} are model-agnostic bounds on the calibration-marginal, input-conditional coverage $P = p(X)$. 
They follow from the Markov and Cantelli inequalities combined with the standard marginal guarantee \eqref{prediction set}.  Corollaries~\ref{corollary lower conditional} and~\ref{corollary cantelli conditional} use~\eqref{calibration conditional} to extend both bounds to the calibration-conditional setup, providing guarantees for $P_D = p_D(X)$.
Theorem~\ref{theorem variance exact} and Corollary~\ref{corollary cantelli conditional} include a tighter, model-dependent Cantelli bound that depends on
$\Var_X(P_D) = \E_X\{P_D - \E_X(P_D)\}^2$.
Assuming an estimator with a uniformly bounded estimation error of $P_D$ is available, Theorem~\ref{theorem cantelli estimated} proves an empirical version of Corollary~\ref{corollary cantelli conditional} for use in real-world scenarios, applicable to methods such as conformal quantile regression \citep{romano2019conformalized}, normalized conformal prediction \citep{papadopoulos2008normalized}, regularized adaptive prediction sets \citep{angelopoulos2020uncertainty} and locally weighted conformal prediction \citep{guan2019conformal}. 
Theorem~\ref{theorem cantelli estimated} also shows that $\Var_X(P_D)$ serves as a computationally feasible surrogate of existing conditional validity estimators such as worst-slab coverage \citep{cauchois2021knowing} and $L_1$-ERT \citep{braun2025conditional}.

\subsection{Related work}

Inductive conformal prediction is widely used to quantify the uncertainty of machine learning prediction systems \citep{vovk2005algorithmic, papadopoulos2008inductive}. 
Input-conditional coverage has been improved empirically by modifying the base predictor \citep{romano2019conformalized}, reweighting calibration samples \citep{vovk2003mondrian, guan2019conformal, lin2021locally}, training adaptive conformity scores \citep{papadopoulos2008normalized, lei2012distribution}, and estimating conditional densities \citep{chernozhukov2021distributional, izbicki2019flexible,
sesia2021conformal}.

Recent work on reducing input-conditional miscoverage has used Lipschitz continuity assumptions to bound $P = p(X)$ \citep{barber2021limits, guan2023localized, gibbs2025conformal, chernozhukov2021distributional}.
To the best of our knowledge, our bounds are the first to avoid such assumptions explicitly.

The impossibility result of \citet{vovk2012conditional} and \citet{lei2012distribution} can be partially bypassed by using the condition $\E_{XYD}[h(X)\{{\bf 1}(Y \in \Gamma_\alpha(X)) - [1-\alpha)\}] = 0$, where $h\colon\cX \to \R$ is any measurable function, as an alternative feasible definition of conditional validity \citep{deng2023happymap, gibbs2025conformal}.
Further options include group-conditional or label-conditional validity \citep{vovk2003mondrian, jung2022batch, ding2023class} and distribution-fitting approaches \citep{einbinder2022training, colombo2024normalizing}.

An increasingly popular line of work focuses on calibration-conditional coverage \citep{vovk2012conditional, bian2023training, duchi2025few}. 
Theorem~2 of \citet{sesia2021conformal} uses a PAC-style bound on conditional validity in a similar spirit, but requires independent and identically distributed data and continuous bounded variables.

Markov's inequality has been used in conformal prediction work based on e-values \citep{bashari2023derandomized, balinsky2024enhancing, vovk2025conformal, gauthier2025values}.  
Cantelli's inequality was exploited in conformal prediction to prove asymptotic optimality of inductive conformal prediction \citep{vovk2002asymptotic} and to test randomness online \citep{vovk2005algorithmic}.

\section{Method}
\label{sec:method}

We focus on regression to simplify presentation. 
Since the bounds do not depend on the label space structure, all results generalize to classification without change.  
We assume equal-size calibration, training, and evaluation datasets $D$, $D_{\rm train}$, and $D_{\rm eval}$ drawn independently and identically from $\P_{XY}$, together with a pre-trained point predictor $f(X) \approx \E(Y \mid X)$.  
Define the bounded random variables
\begin{align}
P &= p(X) = \P_{YD}\{Y \in \Gamma_\alpha(X)\mid X\}, \label{def P} \\
P_D &= p_D(X) = \P_{Y}\{Y \in \Gamma_\alpha(X)\mid X, D\}. \label{def PD}
\end{align}

\subsection{Inductive conformal prediction}
\label{section icp}

For a test sample $(X, Y)$ with $X\in \cX$ and $Y \in \R$, the prediction interval of the point predictor $f(X)$ is
\begin{equation}
\label{prediction interval}
 \Gamma_\alpha(X) = [f(X)-q_\alpha,\; f(X) + q_\alpha],
\end{equation}
where $q_\alpha$ is the $\lceil(1-\alpha)(n+1)\rceil$-th smallest value of $\{|f(X_i)-Y_i|\}_{i=1}^n$.  
For any $\alpha \in (0, 1)$,
\begin{equation}
\label{validity interval}
   \P_{XYD}\{Y\in \Gamma_\alpha(X)\} \geq 1 - \alpha.
\end{equation}
Equivalently, the expectation of $P$ satisfies
\begin{equation}
\label{validity interval expectation}
\E_X(P) \geq 1 - \alpha.
\end{equation}

\subsubsection*{Markov bound}
\label{section markov}

We combine the constraint on $\E_X(P)$ with Markov's inequality to prove the following.

\begin{theorem}[Markov bound]
\label{theorem lower}
For any $\delta, \alpha \in (0, 1)$,
\begin{equation}
\P_X\!\left( P \geq 1-\frac{\alpha}{\delta} \right) \geq 1-\delta.
\end{equation}
\end{theorem}

\begin{proof}[Proof of Theorem~\ref{theorem lower}]
For any $S \in [0, 1]$, Markov's inequality gives $\P(S \geq a) \leq \E(S)/a$ for $a>0$.  
Applying this to $1-S$,
\[
  \P(1-S \geq a) \leq \tfrac{1-\E(S)}{a},
  \quad\text{so}\quad
  \P\bigl(S \geq 1 - a\bigr) \geq 1-\tfrac{1-\E(S)}{a}.
\]
Setting $S = P$ and $\delta = \{1-\E(P)\}/a$ gives
\[
  \P_X\!\left(P \geq 1 - \tfrac{1-\E(P)}{\delta}\right) \geq 1-\delta.
\]
Since $\E_X(P) \geq 1-\alpha$ by \eqref{validity interval expectation}, we have $\alpha/\delta \geq \{1-\E_X(P)\}/\delta$, so the threshold $1-\alpha/\delta \leq 1-\{1-\E_X(P)\}/\delta$ and the result follows.
\end{proof}

The bound, which depends only on the target coverage $\alpha$ and the PAC parameter $\delta$, holds without assumptions on the data distribution and for any valid conformal predictor.  
It does not contradict the impossibility results of \citet{lei2012distribution} and \citet{vovk2012conditional} because it becomes uninformative at high success probabilities.

\subsubsection*{Cantelli bound}
\label{section cantelli}

Replacing Markov's inequality with Cantelli's gives the following result.

\begin{theorem}[Cantelli bound]
    \label{theorem variance exact}
    For any $\delta \in (0, 1)$ and $\alpha < \tfrac12$,
    \begin{align}
      &\P_X\!\left(P \geq 1-\alpha-
        \sqrt{ \tfrac{\alpha(1-\alpha)(1 - \delta)}{\delta}}\right)
        \label{cantelli agnostic}\\
      &\quad\geq
        \P_X\!\left(P \geq 1-\alpha-
        \sqrt{ \tfrac{ \Var_X(P) (1 - \delta)}{\delta}}\right)
        \label{cantelli variance}\\
      &\quad\geq 1 - \delta. \nonumber
    \end{align}
\end{theorem}

\begin{proof}[Proof of Theorem~\ref{theorem variance exact}]
Applying Cantelli's inequality to $1-P$ gives
\[
  \P_X\{-(P - \E_X(P)) \geq t\} \leq \tfrac{\Var_X(P)}{\Var_X(P) + t^2},
\]
using $\E_X(1-P)=1-\E_X(P)$ and $\Var_X(1-P)=\Var_X(P)$.  
Setting $\delta = \Var_X(P)/\{\Var_X(P) + t^2\}$ gives $t = \{\Var_X(P)(1-\delta)/\delta\}^{1/2}$, and hence
\[
    \P_X\!\left(P\geq \E_X(P) -
    \sqrt{\tfrac{\Var_X(P) (1-\delta)}{\delta}}\right) \geq 1 - \delta.
\]
The model-dependent bound follows from $\E_X(P) \geq 1 - \alpha$, implied by \eqref{validity interval expectation}.  
For the model-agnostic bound, the Bhatia--Davis inequality $\Var(S) \leq \{s_+ - \E(S)\}\{\E(S) - s_-\}$ for $S\in[s_-,s_+]$ gives $\Var_X(P) \leq \{1-\E_X(P)\}\E_X(P)$.  
Since $\alpha < \tfrac12$ implies $\E_X(P) \in [1-\alpha, 1] \subseteq (\tfrac12, 1]$ and the function $s(1-s)$ is strictly decreasing on $(\tfrac12, 1]$, the maximum over $\E_X(P) \geq 1-\alpha$ is attained at $\E_X(P) = 1-\alpha$, giving $\Var_X(P) \leq \alpha(1-\alpha)$.
\end{proof}

The factor $\alpha(1-\alpha)$ in \eqref{cantelli agnostic} provides an upper bound on the variance-dependent term in \eqref{cantelli variance}.  
Unlike $\E_X(P)$, which \eqref{validity interval} constrains for every calibration set, $\Var_X(P)$ reflects the underlying conformal predictor's adaptivity. 
For algorithms whose conditional coverage is approximately constant across the attribute space, $\Var_X(P)$ is small.  
A perfectly adaptive algorithm would have $P \to 1-\alpha$ for all $X$, so $\Var_X(P) \to 0$ as $n\to\infty$.
The bound shows that exact conditional validity is asymptotically achievable only in this ideal case.

\subsection{Calibration-conditional bounds}
\label{section icp calibration conditional}

In practice, an inductive conformal prediction model is calibrated once and then evaluated on new samples.  
Hoeffding's inequality implies \citep{vovk2012conditional, duchi2025few}
\begin{equation}
\label{validity interval conditional}
\P_{D}\!\left\{\E_X(P_D) \geq 1-\alpha -\gamma\right\}
\geq 1- e^{-2n\gamma^2},
\end{equation}
where $n=|D|$ and $\alpha, \gamma \in (0, 1)$, using $\P_{XY}(Y\in \Gamma_\alpha \mid D) = \E_X\P_{Y}(Y \in \Gamma_\alpha(X)\mid X, D) = \E_X(P_D)$.
Replacing \eqref{validity interval expectation} with \eqref{validity interval conditional} in Theorem~\ref{theorem lower} gives the following.

\begin{corollary}[Calibration-conditional Markov bound]
\label{corollary lower conditional}
For $\alpha, \gamma, \delta \in (0,1)$ and $n = |D|$,
\begin{equation}
\label{markov for conditional}
\P_{XD}\!\left( P_D \geq 1 - \tfrac{\alpha + \gamma}{\delta}
\right) \geq 1 - \delta  - e^{-2n \gamma^2}.
\end{equation}
\end{corollary}

\begin{proof}[Proof of Corollary~\ref{corollary lower conditional}]
Let $A = \{P_D \leq 1 - \{1-\E_X(P_D)\}/\delta\}$ and $B = \{\E_X(P_D) \leq 1-\alpha-\gamma\}$.  
For any $D$, Markov's inequality and \eqref{validity interval conditional} give $\P_X(A) \leq \delta$ and $\P_D(B) \leq e^{-2n\gamma^2}$.
A union bound gives $\P_{XD}(\lnot A \cap \lnot B) \geq 1 - \delta - e^{-2n\gamma^2}$.
On the intersection, $\E_X(P_D) \geq 1-\alpha-\gamma$ implies $1 - \{1-\E_X(P_D)\}/\delta \geq 1 - (\alpha+\gamma)/\delta$, so $P_D \geq 1 - (\alpha+\gamma)/\delta$, which gives \eqref{markov for conditional}.
\end{proof}

Similarly, we have the following.

\begin{corollary}[Calibration-conditional Cantelli bound]
\label{corollary cantelli conditional}
For any $\alpha, \delta, \gamma \in (0, 1)$ with $\alpha + \gamma < \tfrac12$,
\begin{align}
\label{cantelli agnostic conditional}
        &\P_{XD}\!\left(P_D \geq 1-\alpha-
          \sqrt{\tfrac{(\alpha+ \gamma)(1 - \alpha - \gamma)(1-\delta)}{\delta}}
          - \gamma \right)\\
        &\quad\geq \P_{XD}\!\left(P_D \geq 1-\alpha-
          \sqrt{\tfrac{\Var_X(P_D) (1-\delta)}{\delta}}
          - \gamma \right) \label{cantelli variance conditional}\\
        &\quad\geq 1- \delta - e^{- 2n \gamma^2}. \nonumber
\end{align}
\end{corollary}

\begin{proof}[Proof of Corollary~\ref{corollary cantelli conditional}]
Let $r(\delta)=\{\Var_X(P_D)(1-\delta)/\delta\}^{1/2}$, $A = \{P_D \leq \E_X(P_D)- r(\delta)\}$ and $B = \{\E_X(P_D) \leq 1-\alpha-\gamma\}$.  
For any $D$, Cantelli's inequality and \eqref{validity interval conditional} give $\P_{XD}(A)\leq\delta$ and $\P_D(B) \leq e^{-2n\gamma^2}$.
A union bound gives $\P_{XD}(\lnot A \cap \lnot B) \geq 1 - \delta - e^{-2n\gamma^2}$.
On the intersection, $P_D \geq 1-\alpha-\gamma-r(\delta)$, giving \eqref{cantelli variance conditional}.  
Applying Bhatia--Davis and the monotonicity of $s(1-s)$ on $s\in(\tfrac12,1]$, as in the proof of Theorem~\ref{theorem variance exact}, with $\E_X(P_D)\geq 1-\alpha-\gamma$ yields $\Var_X(P_D)\leq(\alpha+\gamma)(1-\alpha-\gamma)$, which gives \eqref{cantelli agnostic conditional}.
\end{proof}

The variance-dependent bound \eqref{cantelli variance conditional} does not require averaging over all possible calibration sets and can be estimated from data, as described in Section~\ref{section variance estimation}.

\subsection{Variance estimation}
\label{section variance estimation}

The bounds \eqref{cantelli variance} and \eqref{cantelli variance conditional} depend on $\Var_X(P)$ and $\Var_X(P_D)$, which are generally unknown.  
Replacing $\Var_X(P_D)$ by an empirical estimate introduces estimation and finite-sample errors.
Assuming the conditional variance estimator has bounded errors, we prove the following. 

\begin{theorem}
\label{theorem cantelli estimated}
Let $\hat p(X)$ be an estimator of $p_D(X)$ satisfying $\hat p(X) = p_D(X) + \varepsilon$, $|\varepsilon| \leq \delta_\varepsilon$, and let $D_{\rm eval}$, with $|D_{\rm eval}| = n$, be an evaluation sample independent of $D$ and of the training data used to fit $\hat p$.
Define
\begin{align}
\label{empirical mean and variance}
  \hat\tau &= \tfrac{1}{n}\sum_{i=1}^n \hat p(X_i),
  &
  \hat V   &= \tfrac{1}{n-1}\sum_{i=1}^n \bigl\{\hat p(X_i)-\hat\tau\bigr\}^2,
  \quad X_i \in D_{\rm eval}.
\end{align}
Then, for any $t > 0$, with probability at least $1 - 2e^{-2nt^2}$ over $D_{\rm eval}$,
\begin{equation}
\label{cantelli variance estimated final}
    \P_{XD}\!\left(P_D \geq 1-\alpha-\Delta\right) \geq 1-\delta - e^{-2n\gamma^2},
\end{equation}
where $\Delta = \{(\hat V + 2\delta_\varepsilon + \delta_\varepsilon^2 + 3t)(1-\delta)/\delta\}^{1/2} + \gamma$.
\end{theorem}

\begin{proof}[Proof of Theorem~\ref{theorem cantelli estimated}]
Since $|\varepsilon|\leq\delta_\varepsilon$, the triangle inequality for standard deviations gives $\Var_X(p_D(X))^{1/2} \leq \Var_X(\hat p(X))^{1/2} + \Var(\varepsilon)^{1/2} \leq \Var_X(\hat p(X))^{1/2} + \delta_\varepsilon$.
Using $\Var_X(\hat p(X))^{1/2} \leq 1$,
\begin{equation}
\label{var bound step1}
  \Var_X(p_D(X)) \leq
  \bigl\{\Var_X(\hat p(X))^{1/2}+\delta_\varepsilon\bigr\}^2 \leq \Var_X(\hat p(X)) + 2\delta_\varepsilon,
\end{equation}
where we used $2\sqrt{\Var_X(\hat p(X))}\delta_\varepsilon  + \delta_\varepsilon^2\leq 2 \delta_\varepsilon$, implied by Bhatia–Davis inequality, $\sqrt{\Var_X(\hat p_D(X))} \leq \frac12$, and $\hat p(X) \in [0, 1]$, $\delta^2_\varepsilon \leq \delta_\varepsilon$.
Inserting \eqref{var bound step1} into Corollary~\ref{corollary cantelli conditional} gives 
\begin{align}
\label{cantelli variance estimated}
    \P_{XD}\!\biggl(P_D \geq 1-\alpha-
      \biggl\{\frac{(\Var_X(\hat p(X))+2\delta_\varepsilon)(1-\delta)}
                  {\delta}\biggr\}^{1/2}
      - \gamma \biggr)
    \geq 1-\delta - e^{-2n\gamma^2}.
\end{align}
To bound $\Var_X(\hat p(X))$ in terms of \eqref{empirical mean and variance},  write $\Var_X(\hat p(X)) = \mu_2 - \mu_1^2$, where $\mu_2 = \E_X\{\hat p(X)^2\}$ and $\mu_1 = \E_X\{\hat p(X)\}$, and let $\bar{m}_2 = n^{-1}\sum_{i=1}^n \hat p(X_i)^2$ and
$\bar{m}_1 = \hat\tau$.  
Since $\hat p(X_i)\in[0,1]$, both $\hat p(X_i)$ and $\hat p(X_i)^2$ lie in $[0,1]$, so the one-sided Hoeffding inequality gives
\[
  \P(\mu_2 > \bar{m}_2 + t) \leq e^{-2nt^2},
  \qquad
  \P(\mu_1 < \bar{m}_1 - t) \leq e^{-2nt^2}.
\]
A union bound yields, with probability at least $1-2e^{-2nt^2}$,
\[
  \Var_X(\hat p) = \mu_2 - \mu_1^2
  \leq (\bar{m}_2 + t) - (\bar{m}_1 - t)^2.
\]
Expanding and using $\bar{m}_2 - \bar{m}_1^2 = \{(n-1)/n\}\hat V \leq \hat V$ and $\bar{m}_1\leq 1$,
\begin{equation}
\label{var bound step2}
  \Var_X(\hat p(X)) \leq \hat V + t(1 + 2\bar{m}_1) - t^2
                 \leq \hat V + 3t.
\end{equation}
Combining \eqref{var bound step1} and \eqref{var bound step2} gives $\Var_X(p_D) \leq \hat V + 2\delta_\varepsilon + 3t$, which holds with probability at least $1 - 2e^{-2nt^2}$.  
Substituting into \eqref{cantelli variance estimated} and using the monotonicity of the bound in $\Var_X(P_D)$ gives \eqref{cantelli variance estimated final}. 
\end{proof}

The parameter $t$ is free and can be interpreted as an additional PAC parameter.    
Given $\delta_\varepsilon$, the bound is fully non-asymptotic and distribution-free for all $n$, without requiring the conditional validity estimator $\varepsilon$ to be unbiased.
In practice, we use a calibrated conformal predictor and a $k$-nearest-neighbour estimator to transform $D_{\rm train}$ into
attribute--coverage pairs $(X_i, Z_i)$ with $Z_i = {\bf 1}(Y_i \in \Gamma_\alpha(X_i))$, and approximate $p_{D}(X)$ by
\begin{equation}
\label{knn estimator}
  \phat(X) = k_n^{-1}\sum_{(X', Y') \in N_{k_n}(X)}
             {\bf 1}(Y' \in \Gamma_\alpha(X')) \approx p_D(X),
\end{equation}
where $N_{k_n}(X)\subseteq D_{\rm train}$ denotes the $k_n$ nearest neighbours of $X$ and $k_n = \lfloor\log n\rfloor$.  
The choice $k_n = \log n$ ensures $k_n \to\infty$ and $k_n/n\to 0$, which imply the asymptotic convergence of $\phat$ when the target is bounded \citep{devroye1994strong}.
The corresponding estimated mean and variance on $D_{\rm eval}$ are as in \eqref{empirical mean and variance}.
Theorem~\ref{theorem cantelli estimated} does not require this specific estimator. Any other bounded estimator may be used.

\subsection{Measuring conditional validity}
To the best of our knowledge, no existing work explicitly estimates $\Var_X(P_D)$. 
Existing conditional validity estimators focus on related quantities, including worst-slab coverage and $L_1\text{-ERT} = \E_X(|P-(1-\alpha)|)$.
One has $\Var_X(P) \to L_2\text{-ERT} = \E_X\{P-(1-\alpha)\}^2$ as $n\to\infty$, since $\E_X(P_n)\to 1-\alpha$ implies $\lim_{n\to\infty}\Var_X(P_n) = \lim_{n\to\infty}\E_X\{P_n-(1-\alpha)\}^2$.
Worst-slab coverage fits in the ERT framework through $\ell(\beta) =\E_X\{P e^{-\beta P}/\E_X(e^{-\beta P})\} \approx \mathrm{ess}\inf_{X} P$ for $\beta\gg 1$ \citep{braun2025conditional}.
Table~\ref{table measures} summarizes the ERT framework.

\begin{table}[t]
    \centering
    \caption{The proposed $k$-nearest-neighbour variance measure and
    worst-slab coverage in the $\ell$-ERT framework
    \citep{braun2025conditional}.
    For all measures, $\hat p_i = \hat p(X_i) \approx p(X_i)$ with
    $\hat p(X)$ as in~\eqref{knn estimator}, and
    $\hat\tau = n^{-1}\sum_{i=1}^n \hat p_i \to 1-\alpha$ as
    $|D|\to\infty$.}
    \begin{tabular}{ccc}
    \toprule
    $\ell$ & $n\to \infty$ & $k$-nearest-neighbour estimate \\
    \midrule
    $L_1$ & $\E_X(|P - (1-\alpha)|)$     & $n^{-1}\|\hat p - \hat\tau\|_1$ \\
    $L_2$ & $\E_X\{P - (1-\alpha)\}^2$   & $n^{-1}\|\hat p - \hat\tau\|_2^2$ \\
    $L_{-\infty}$ & $\mathrm{ess}\inf_X P$ & $\min_i \hat p_i$ \\
    \bottomrule
    \end{tabular}
    \label{table measures}
\end{table}

\section{Empirical validation}
\label{sec:experiments}

\subsection{Data and methods}

We evaluate all bounds and conditional coverage proxies on four classification and regression datasets: CIFAR-10 \citep{cifar10}, Forest Cover Type \citep{covertype}, Communities and Crime \citep{communities_and_crime}, and Airfoil Self-Noise \citep{airfoil}.  
Available samples are split into five subsets with proportions $(\frac{1}{2}, \frac{1}{8}, \frac{1}{8}, \frac{1}{8}, \frac{1}{8})$ for fitting, calibrating, training the variance estimator, estimating coverage variance, and testing.  
We pre-encode CIFAR-10 images into an 80-dimensional feature space using a pretrained VGG network without the classification head.

We consider four conformal prediction approaches.
Inductive conformal prediction is the baseline from
Section~\ref{section icp}.
Conformal quantile regression \citep{romano2019conformalized} estimates conformally calibrated lower and upper conditional quantiles.
For classification, it is replaced by regularized adaptive prediction sets \citep{angelopoulos2020uncertainty}.
Normalized inductive conformal prediction \citep{papadopoulos2008normalized} accounts for heteroskedasticity by scaling conformity scores with a fitted estimate of local variability.
Locally weighted conformal prediction \citep{guan2019conformal} is a kernel-based reweighting method that adapts prediction intervals by up-weighting nearby calibration points.

Table~\ref{table performance} shows performance on all datasets.
\begin{table}[t]
  \centering
  \caption{Performance of four conformal prediction methods on all datasets with $\alpha=0.05$.  Entries show mean $\pm$ standard deviation over 10 random splits.}
  \begin{tabular}{lcccc}
  \toprule
  Method & Marginal cov. & Interval size & $\widehat{\Var}$ cond.\ cov. & Mean cond.\ cov. \\
  \midrule
  \multicolumn{5}{l}{\textit{Communities and Crime}} \\
  CQR         & $0.958 \pm 0.012$ & $0.683 \pm 0.059$ & $0.007 \pm 0.003$ & $0.969 \pm 0.013$ \\
  ICP         & $0.952 \pm 0.012$ & $0.645 \pm 0.063$ & $0.009 \pm 0.003$ & $0.963 \pm 0.014$ \\
  LWCP        & $0.939 \pm 0.014$ & $0.525 \pm 0.040$ & $0.010 \pm 0.003$ & $0.950 \pm 0.012$ \\
  Normalized  & $0.947 \pm 0.015$ & $0.547 \pm 0.054$ & $0.009 \pm 0.004$ & $0.956 \pm 0.022$ \\
  \midrule
  \multicolumn{5}{l}{\textit{Airfoil}} \\
  CQR         & $0.947 \pm 0.015$ & $21.013 \pm 0.728$ & $0.015 \pm 0.005$ & $0.943 \pm 0.018$ \\
  ICP         & $0.951 \pm 0.011$ & $14.180 \pm 0.656$ & $0.013 \pm 0.006$ & $0.949 \pm 0.017$ \\
  LWCP        & $0.941 \pm 0.020$ & $13.326 \pm 0.514$ & $0.013 \pm 0.004$ & $0.942 \pm 0.015$ \\
  Normalized  & $0.953 \pm 0.018$ & $14.264 \pm 0.745$ & $0.012 \pm 0.005$ & $0.950 \pm 0.019$ \\
  \midrule
  \multicolumn{5}{l}{\textit{Forest Cover Type}} \\
  ICP         & $0.945 \pm 0.005$ & $1.855 \pm 0.057$ & $0.010 \pm 0.002$ & $0.952 \pm 0.007$ \\
  LWCP        & $0.938 \pm 0.005$ & $1.952 \pm 0.079$ & $0.009 \pm 0.001$ & $0.945 \pm 0.009$ \\
  Normalized  & $0.948 \pm 0.008$ & $3.782 \pm 0.427$ & $0.009 \pm 0.002$ & $0.950 \pm 0.008$ \\
  RAPS        & $0.947 \pm 0.007$ & $2.059 \pm 0.037$ & $0.008 \pm 0.001$ & $0.948 \pm 0.007$ \\
  \midrule
  \multicolumn{5}{l}{\textit{CIFAR-10}} \\
  ICP         & $0.947 \pm 0.004$ & $5.147 \pm 0.133$ & $0.007 \pm 0.001$ & $0.952 \pm 0.007$ \\
  LWCP        & $0.930 \pm 0.006$ & $4.940 \pm 0.121$ & $0.008 \pm 0.001$ & $0.941 \pm 0.009$ \\
  Normalized  & $0.947 \pm 0.008$ & $7.449 \pm 0.159$ & $0.007 \pm 0.001$ & $0.954 \pm 0.005$ \\
  RAPS        & $0.949 \pm 0.005$ & $5.383 \pm 0.125$ & $0.007 \pm 0.001$ & $0.952 \pm 0.006$ \\
  \bottomrule
  \end{tabular}
  \label{table performance}
\end{table}

\subsection{Heteroskedasticity}

\begin{figure}
    \centering
    \includegraphics[width=0.9\linewidth]{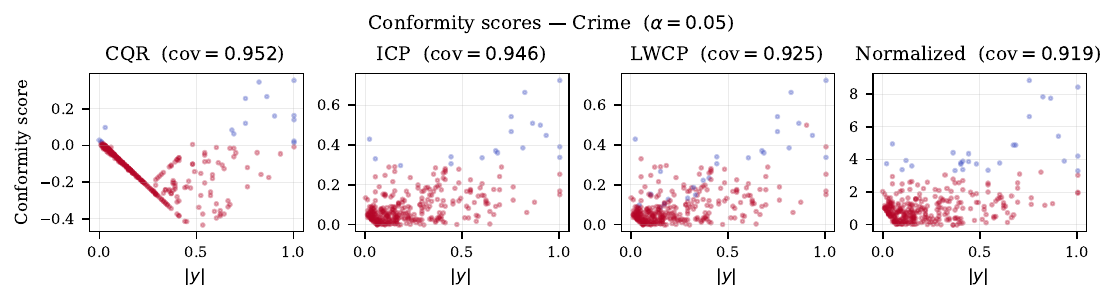}
    \includegraphics[width=0.9\linewidth]{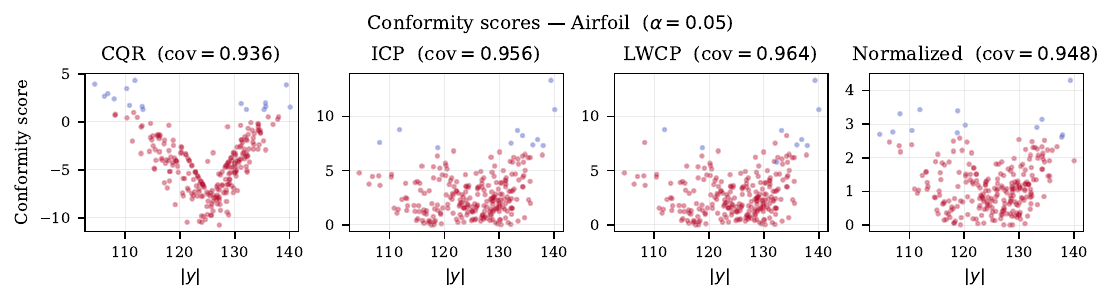}
    \includegraphics[width=0.9\linewidth]{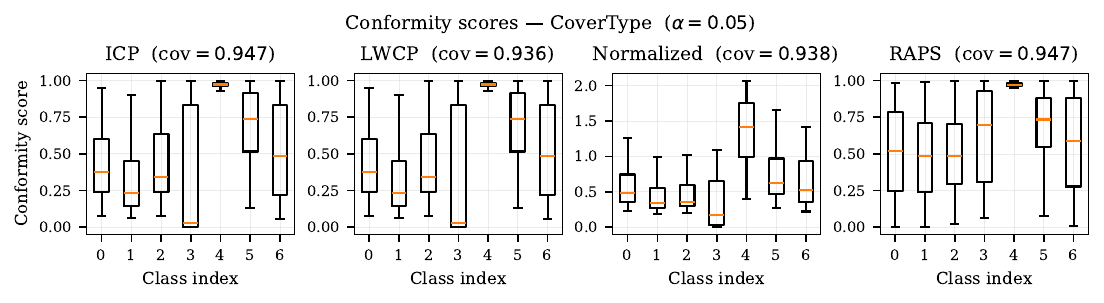}
    \includegraphics[width=0.9\linewidth]{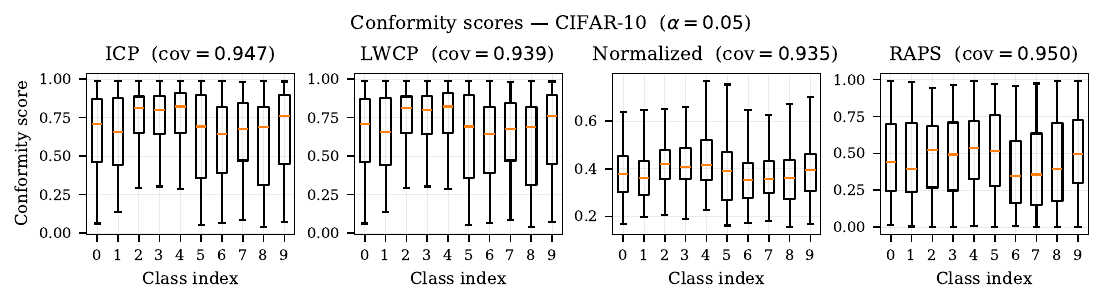}
    \caption{Conformity scores plotted against the target value for each
    of the four datasets.}
    \label{figure residuals}
\end{figure}

We visualise heteroskedasticity by plotting conformity scores against true responses (Fig.~\ref{figure residuals}).  
For regression, scatter plots use different markers for covered and uncovered instances. 
For classification, per-class boxplots show class-conditional statistics for each method's conformity scores.

\subsection{Lower bounds comparison}
Table~\ref{table parametric bounds} summarises all bounds as a function of the PAC parameters, $\alpha$ and $\delta$.
\begin{table}[t]
  \centering
  \begin{tabular}{lcc}
  \toprule
    Bound & Coverage lower bound & Success probability \\
  \midrule
    Markov        & $1 - \alpha/\delta$
                  & $1 - \delta$ \\
    Cantelli      & $1 - \alpha - \{\alpha(1-\alpha)(1-\delta)/\delta\}^{1/2}$
                  & $1 - \delta$ \\
    Markov-cond   & $1 - (\alpha + \gamma)/\delta$
                  & $1 - \delta - e^{-2n\gamma^2}$ \\
    Cantelli-cond & $1 - \alpha - \{\alpha(1-\alpha)(1-\delta)/\delta\}^{1/2} - \gamma$
                  & $1 - \delta - e^{-2n\gamma^2}$ \\
    Cantelli-est  & $1 - \alpha - \{\hat V(1-\delta)/\delta\}^{1/2} - \gamma$
                  & $1 - \delta - e^{-2n\gamma^2}$ \\
  \bottomrule
  \end{tabular}
    \caption{Parametrized bounds used in the experiments.  In all simulations, we set $\gamma=0.1$ and assume the variance estimator is correct ($\delta_\varepsilon=0$ and $n \to \infty$ in Theorem~\ref{theorem cantelli estimated}).}
  \label{table parametric bounds}
\end{table}

\begin{figure}
  \begin{center}
    \includegraphics[width=.48\textwidth]{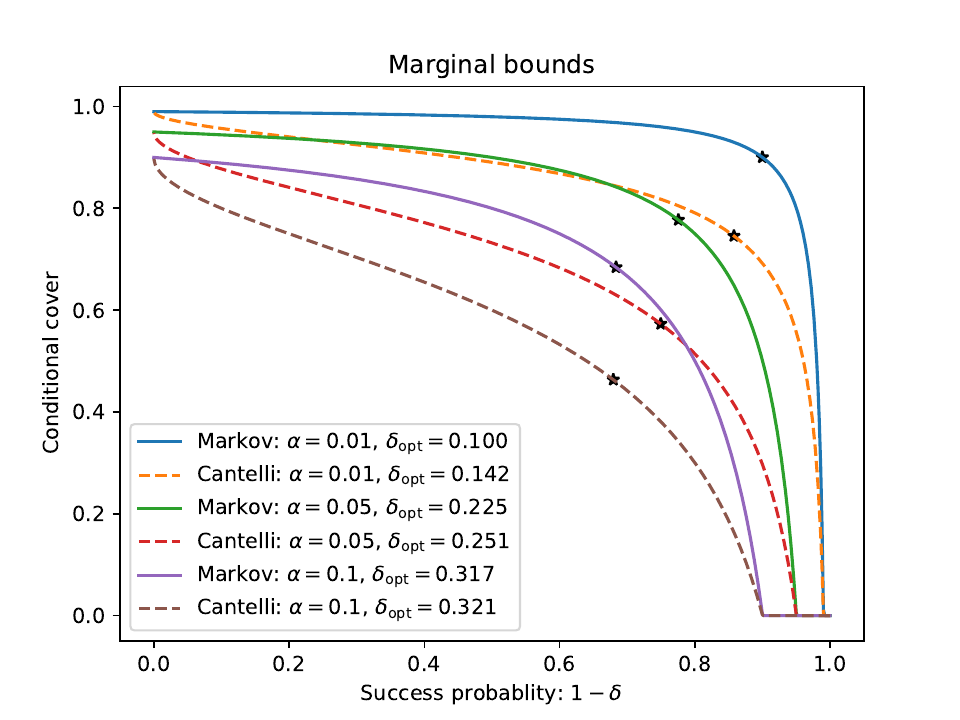}\hfill
    \includegraphics[width=.48\textwidth]{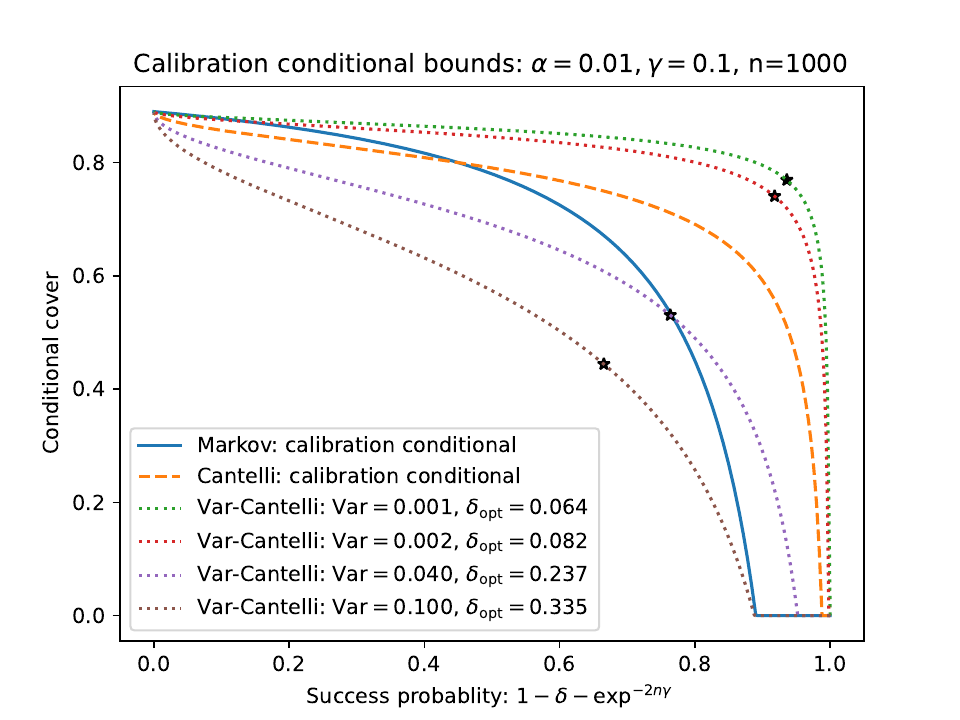}
  \end{center}
  \caption{Left: guaranteed conditional coverage for different target coverages as a function of the success probability.  Right: variance-dependent bounds for different values of $\Var_X(p_D(X)) = \E_X\{p_D(X)-\E_X p_D(X)\}^2$, where $p_D(X) = \P_{Y}\{Y \in \Gamma_\alpha(X)\mid X, D\}$.
  Stars indicate the PAC-optimal choices of $\delta$ for given $\alpha$ that maximize the union bound between the inner and outer probabilities.
  See Table~\ref{table parametric bounds} for parametric forms of all bounds.}
  \label{figure model agnostic bounds}
\end{figure}

Figure~\ref{figure model agnostic bounds} compares the coverage--success profile established by the distribution-free bounds for different values of $\alpha$ (left panel) and shows the gap between the model-agnostic lower bounds and their variance-dependent versions for several values of $\Var_X(P_D)$ (right panel).

We use $n$ attribute--coverage pairs obtained from the evaluation set to train the variance estimator, compute $\hat V \approx \Var_X(P_D)$, and plot the Cantelli-conditional bound for each conformal predictor. 
\begin{figure}
    \centering
    \includegraphics[width=0.9\linewidth]{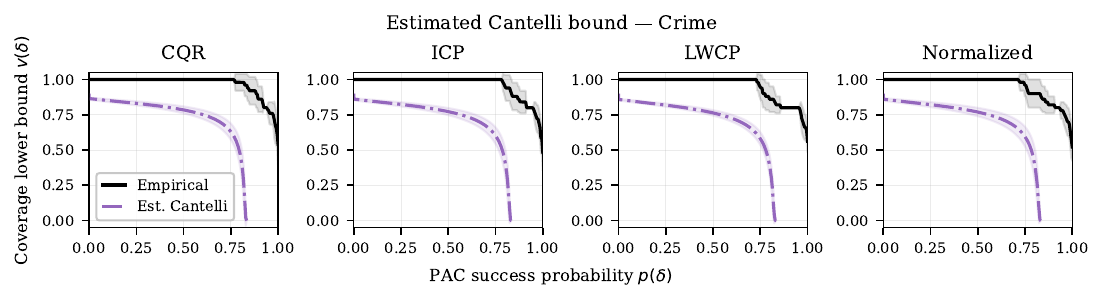}
    \includegraphics[width=0.9\linewidth]{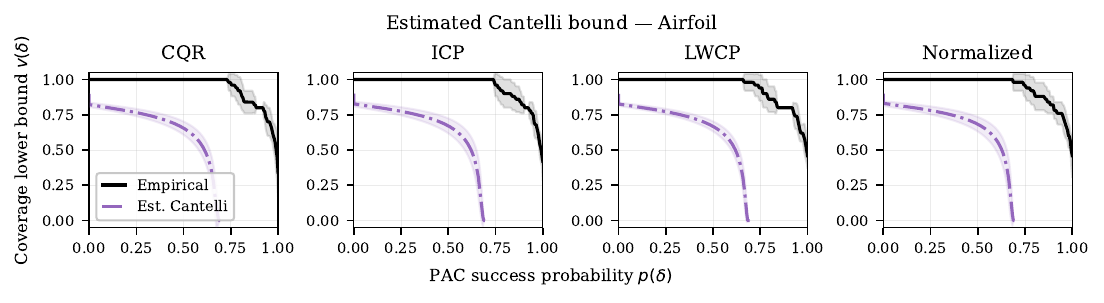}
    \includegraphics[width=0.9\linewidth]{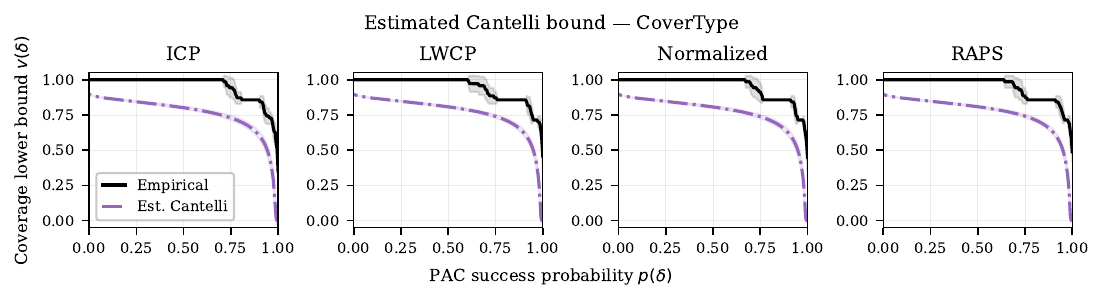}
    \includegraphics[width=0.9\linewidth]{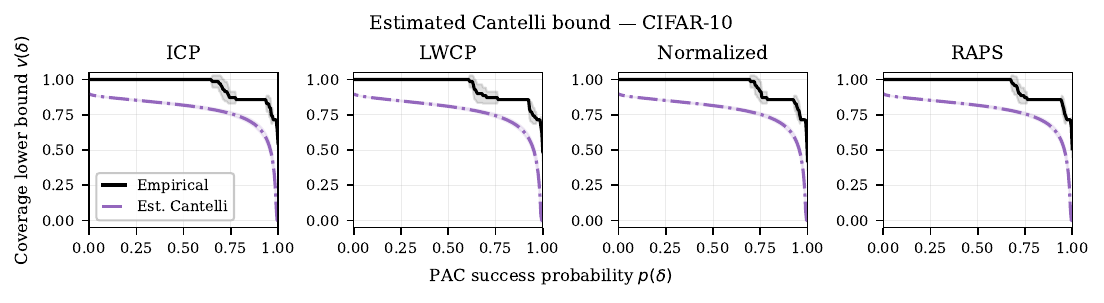}
    \caption{Conditional coverage bounds and empirical estimates for all four datasets.  
    Each panel shows the estimated Cantelli-est bound from Table~\ref{table parametric bounds} as a function of the
    success probability.  
    Solid lines are the test empirical coverage, obtained by estimating $\hat p(X)$ on the test set and plotting the
    $(1-\delta)$-quantile.  
    Curves are averages over ten random splits. Shaded regions show $\pm 1$ standard deviation.} 
    \label{figure bounds}
\end{figure}

\subsection{Conditional coverage estimators}

We compare worst-slab coverage and $L_1$-ERT \citep{cauchois2021knowing, braun2025conditional} with estimates of $\Var_X(P_D)$ obtained via~\eqref{knn estimator}.  
We train the estimators on random subsets of $D_{\rm train} \cup D_{\rm eval}$ of increasing sizes $m$ to assess robustness and data efficiency. 
We only use ICP prediction sets in this experiment.

\begin{figure}
    \centering
    \includegraphics[width=.99\linewidth]{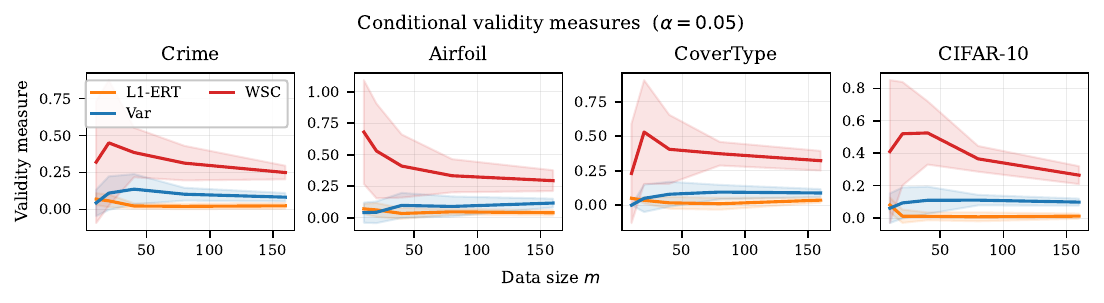}
    \caption{Comparison of $\hat V \approx \Var_X(P_D)$ with $L_1$-ERT and worst-slab coverage from Table~\ref{table measures}, for increasing sizes of the estimation set.  
    The $x$-axis is $m = |D_{\rm eval}|$, quantities are averaged over ten splits, and shaded regions show $\pm 1$ standard deviation.}
    \label{figure K}
\end{figure}

\section{Discussion}
\label{sec:discussion}

Theorem~\ref{theorem cantelli estimated} requires a bound on the
estimation error $\varepsilon$.  
Such a bound cannot be guaranteed non-asymptotically without further assumptions on finite training sets.
\citet{devroye1994strong} shows the asymptotic consistency of the $k$-nearest-neighbour estimator without assuming Lipschitz continuity of $p_D(X)$.
We deliberately avoided asymptotic guarantees, which may look incompatible with the finite-sample, distribution-free character of conformal inference.  
The proposed model-agnostic bounds are distribution-free and hold for finite calibration samples, even if they are possibly loose.  
The empirical bound of Theorem~\ref{theorem cantelli estimated} isolates the estimation error in quantifying a model's conditional miscoverage. 
One limitation is that an additional dataset $D_{\rm train}$ is needed to estimate $\Var_X(P)$.  
Reserving part of the training data for calibration refinement is common in adaptive conformal prediction algorithms such as normalized inductive conformal prediction \citep{papadopoulos2008normalized}, and our experiments show that the variance estimate remains stable even for relatively small $D_{\rm train}$.   
This estimation dataset does not need to be fully labelled, since the $k$-nearest-neighbour estimator uses only attribute--coverage pairs. 
Exploiting this in real-world applications is left for future work.

\bibliographystyle{abbrvnat}
\bibliography{refs}

\end{document}